\documentclass[a4paper]{llncs}

\usepackage[T1]{fontenc}
\usepackage[utf8]{inputenc}
\usepackage{lmodern}
\usepackage[scaled=.8]{beramono}
\usepackage{latexsym}
\usepackage{textcomp}
\usepackage{algorithm}
\usepackage{algpseudocode}
\usepackage{amsmath}
\usepackage{verbatim}
\usepackage{amssymb}
\usepackage{graphicx}
\usepackage{subfig}
\usepackage{paralist}
\usepackage{url}
\usepackage{comment}
\usepackage{proof}
\usepackage[hidelinks]{hyperref}

\usepackage[final]{listings}
\usepackage{multirow}
\usepackage[textsize=tiny]{todonotes}
\usepackage{pl}
\usepackage{times} % added to get more space

\usepackage{array,multirow,graphicx}

\usepackage{listings}

\usepackage{array}
\newcolumntype{L}{>{\centering\arraybackslash}m{3cm}}
\newcommand{\furl}[1]{\footnote{\scriptsize \url{#1}}} 

\usepackage{color}
\definecolor{gray}{rgb}{0.4,0.4,0.4}
\definecolor{darkblue}{rgb}{0.0,0.0,0.6}
\definecolor{cyan}{rgb}{0.0,0.6,0.6}
\lstdefinestyle{sparql}{%
    morekeywords={SELECT,OPTIONAL,FROM,DISTINCT,a,WHERE,FILTER,GROUP,ORDER,LIMIT,BY,IN,AS},
    emph={r,pub,aairObject,verb,person,bday,s,p,o}
}
\lstdefinestyle{turtle}{%
    morekeywords={a, @prefix},
    morecomment=[s][\textrm]{<}{>},
    morecomment=[s][\textit]{"}{"},
}
\lstdefinestyle{xml}{%
    morekeywords={a, @prefix},
    morecomment=[s][\textrm]{<}{>},
    morecomment=[s][\textit]{"}{"},
    keywordstyle=\color{cyan},
}

\lstdefinelanguage{XML}
{
  stringstyle=\color{black},
  identifierstyle=\color{darkblue},
  keywordstyle=\color{cyan},
}

\begin{document}
	
\mainmatter  % start of an individual contribution
	
% first the title is needed
\title{Unveiling Relations in the Industry 4.0 Standards Landscape based on Knowledge Graph Embeddings}
\titlerunning{Unveiling Relations in the I4.0 Standards Landscape}

\author{ 
Ariam Rivas\inst{1},
Irl\'an Grangel-Gonz\'alez\inst{2}, 
Diego Collarana\inst{3},
Jens Lehmann\inst{3}, \\
Maria-Esther Vidal\inst{1,4}
}
\authorrunning{Grangel-Gonz\'alez et al.}

\institute{%
L3S, Leibniz Univ. of Hannover
   \\
\email{ariam.rivas@tib.eu}
\and
Robert Bosch Corporate Research GmbH, Renningen, Germany\\
\email{irlan.grangelgonzalez@de.bosch.com}
\and 
University of Bonn \& Fraunhofer IAIS \\ 
\email{diego.collarana.vargas@}
\email{\{diego.collarana.vargas|jens.lehmann\}@iais.fraunhofer.de}
\and
TIB Leibniz Information Centre for Science and Technology \\
\email{maria.vidal@tib.eu}
}
	
\toctitle{Unveiling Relations in the Industry 4.0 Standards Landscape based on Knowledge Graph Embeddings}
\maketitle

\begin{abstract}
Industry~4.0 (I4.0) standards and standardization frameworks have been proposed with the goal of \emph{empowering interoperability} in smart factories.
These standards enable the description and interaction of the main components, systems, and processes inside of a smart factory.
Due to the growing number of frameworks and standards, there is an increasing need for approaches that automatically analyze the landscape of I4.0 standards. 
Standardization frameworks classify standards according to their functions into layers and dimensions. However, similar standards can be classified differently across the frameworks, producing, thus, interoperability conflicts among them.
Semantic-based approaches that rely on ontologies and knowledge graphs, have been proposed to represent standards, known relations among them, as well as their classification according to existing frameworks.
Albeit informative, the structured modeling of the I4.0 landscape only provides the foundations for detecting interoperability issues. Thus, graph-based analytical methods able to exploit knowledge encoded by these approaches, are required to uncover alignments among standards.
We study the relatedness among standards and frameworks based on community analysis to discover knowledge that helps to cope with interoperability conflicts between standards.
We use knowledge graph embeddings to automatically create these communities exploiting the meaning of the existing relationships.
In particular, we focus on the identification of similar standards, i.e., communities of standards, and analyze their properties to detect unknown relations.
We empirically evaluate our approach on a knowledge graph of I4.0 standards using the Trans$^*$ family of embedding models for knowledge graph entities.
Our results are promising and suggest that relations among standards can be detected accurately.
\end{abstract}

\section{Introduction}
The international community recognizes Industry~4.0 (I4.0) as the fourth industrial revolution. 
The main objective of I4.0 is the creation of \emph{Smart Factories} by combining the Internet of Things (IoT), Internet of Services (IoS), and Cyber-Physical Systems (CPS).
In smart factories, humans, machines, materials, and CPS need to communicate intelligently in order to produce individualized products.
To tackled the problem of interoperability, different industrial communities have created standardization frameworks. 
Relevant examples are the Reference Architecture for Industry~4.0 (RAMI4.0)~\cite{ramiShellStructure} or the Industrial Internet Connectivity Framework (IICF) in the US~\cite{IIRA2017}.
Standardization frameworks classify, and align industrial standards according to their functions. 
While being expressive to categorize existing standards, standardization frameworks may present divergent interpretations of the same standard. 
%How are we using embeddings to do this analysis
Mismatches among standard classifications generate semantic interoperability conflicts that negatively impact on the effectiveness of communication in smart factories.

%Here we could write the WHAT. Build standards communities that support expert analysis to classifier the standards in their respective layers.
Database and Semantic web communities have extensively studied the problem of data integration ~\cite{GolshanHMT17,kovalenko2016semantic,Mountantonakis:2019:LSI:3362097.3345551}, and various approaches have been proposed to support data-driven pipelines to transform industrial data into actionable knowledge in smart factories~\cite{HodgesGR17,Donovan2015}. Ontology-based approaches have also contributed to create a shared understanding of the domain~\cite{Lelli19}, and specifically  
Kovalenko and Euzenat~\cite{kovalenko2016semantic} have equipped data integration with diverse methods for ontology alignment. Furthermore, Lin \emph{et al.}~\cite{ramiirareport2017} identify interoperability conflicts across domain specific standards (e.g., RAMI4.0 model and the IICF architecture), while works by Grangel-Gonzalez \emph{et al.} ~\cite{etfa-grangel2017,alligatorekaw2016,Grangel-Gonzalez18} show the relevant role that Descriptive Logic, Datalog, and Probabilistic Soft Logic play in liaising I4.0 standards. 
Certainly, the extensive literature in data integration provides the foundations for enabling the semantic description and alignment of "similar" things in a smart factory. Nevertheless, finding alignments across I4.0 requires the encoding of domain specific knowledge represented in standards of diverse nature and standardization frameworks defined with different industrial goals. We rely on state-of-the-art knowledge representation and discovery approaches to embed meaningful associations and features of the I4.0 landscape, to enable interoperability.

We propose a knowledge-driven approach first to represent standards, known relations among them, as well as their classification according to existing frameworks.
Then, we utilize the represented relations to build a latent representation of standards, i.e., embeddings. 
Values of similarity metrics between embeddings are used in conjunction with state-of-the-art community detection algorithms to identify patterns among standards. 
Our approach determines relatedness among standards by computing communities of standards and analyzing their properties to detect unknown relations.
Finally, the {\it homophily} prediction principle is performed in each community to discover new links between standards and frameworks.
We asses the performance of the proposed approach in a data set of 249 I4.0 standards connected by 736 relations extracted from the literature. The observed results suggest that encoding knowledge enables for the discovery of meaningful associations. Our contributions are as follows:
\begin{enumerate}
    \item We formalize the problem of finding relations among I4.0 standards and present $\textit{I4.0}\cal{RD}$, a knowledge-driven approach to unveil these relations. 
    $\textit{I4.0}\cal{RD}$ exploits the semantic description encoded in a knowledge graph via the creation of embeddings, to identify then communities of standards that should be related.
    \item We evaluate the performance of $\textit{I4.0}\cal{RD}$ in different embeddings learning models and community detection algorithms.
    The evaluation material is available \furl{https://github.com/i40-Tools/I40KG-Embeddings}.
\end{enumerate}

The rest of this paper is organized as follows: Section~\ref{sec:motiv} illustrates the interoperability problem presented in this paper.  
 Section~\ref{sec:approach} presents the proposed approach, while the architecture of the proposed solution is explained in Section~\ref{sec:architecture}. Results of the empirical evaluation of our methods are reported in Section~\ref{sec:evaluation} while Section~\ref{sec:relatedwork} summarizes the the state of the art. Finally, we close with the conclusion and future work in section~\ref{sec:conclusion}.
 
% To review and use figure - https://www.iiconsortium.org/pdf/JTG2_Whitepaper_final_20171205.pdf
\begin{figure*}[t!]
  \centering    
  \includegraphics[width=0.85\linewidth]{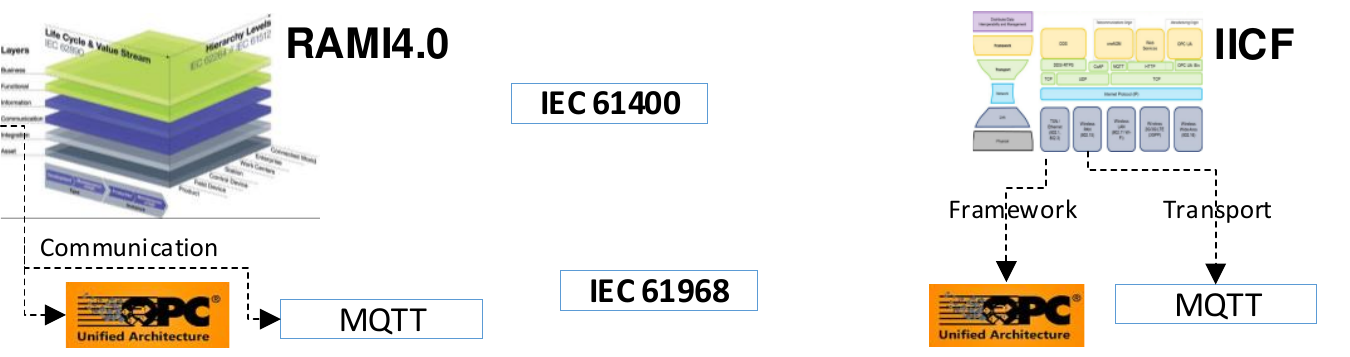}
  \caption{\textbf{Motivating Example}. The RAMI4.0 and IICF standardization frameworks are developed for diverse industrial goals; they classify standards in layers according to their functions, e.g., OPC UA and MQTT under the communication layer in RAMI4.0, and OPC UA and MQTT in the framework and transport layers in IICF, respectively. Further,
 some standards, e.g., IEC 61400 and IEC 61968, are not classified yet.} 
  \label{fig:motivation}
\end{figure*}

\section{Motivating Example}
\label{sec:motiv}
Existing efforts to achieve interoperability in I4.0, mainly focus on the definition of standardization frameworks.
A standardization framework defines different layers to group related I4.0 standards based on their functions and main characteristics.
Typically, classifying existing standards in a certain layer is not a trivial task and it is influenced by the point of view of the community that developed the framework.
RAMI4.0 and IICF are exemplar frameworks, the former is developed in Germany while the latter in the US; they meet specific I4.0 requirements of certain locations around the globe. 
RAMI4.0 classifies the standards OPC UA and MQTT into the Communication layer, stating this, that both standards are similar. Contrary, IICF presents OPC UA and MQTT at distinct layers, i.e., the framework and the transport layers, respectively.
Furthermore, independently of the classification of the standards made by standardization frameworks, standards have relations based on their functions.
Therefore, IEC 61400 and IEC 61968 that are usually utilized to describe electrical features, are not classified at all. 
Figure~\ref{fig:motivation} depicts these relations across the frameworks RAMI4.0 and IICF, and the standards; it illustrates interoperability issues in the I4.0 landscape.

%The function of these standards is enabling the description and the information exchange between I4.0 entities. 
%Most of the standards that are classified by frameworks exist before the frameworks.
%Thus, the number of existing standards that are related to I4.0 overpass those that are classified in the frameworks. 
%Standardization frameworks are developed regionally representing a view of a community w.r.t. I4.0. Therefore, the classifications of the standards in the frameworks are not enough to cope with the demand in I4.0.
%There are two reasons for this:  1) standards that are not classified by the standardization frameworks; 2) standards that are classified differently in distinct standardization frameworks.  
%This is the case of the Open Platform Communications Unified Architecture (OPC UA) and MQTT. Both are standards for performing machine to machine communication. OPC UA and MQTT are categorized in RAMI4.0 together in the Communication layer, whereas in IICF, OPC UA is categorized in the framework layer and MQTT in the transport layer.  Thus, the same standards, i.e., OPC UA and MQTT, belong to two different layers. 
%In addition to that, some standards that can be of relevance for I4.0, are not classified in any of the frameworks, e.g., IEC 61400, IEC 61968, and ISO 20140. 
 
Existing data integration approaches rely on the description of the characteristics of entities to solve interoperability by discovering alignments among them. Specifically, in the context of I4.0, semantic-based approaches have been proposed to represent standards, known relations among them, as well as their classification according to existing frameworks~\cite{baderGTL19,chungoora2013towards,ramiirareport2017,LuMF15}.
Despite informative, the structured modeling of the I4.0 landscape only provides the foundations for detecting interoperability issues.

We propose  $\textit{I4.0}\cal{RD}$, an approach capable of discovering relation over I4.0 knowledge graphs to identify unknown relations among standards. Our proposed methods exploit relations represented in an I4.0 knowledge graph to compute the similarity of the modeled standards. Then, an unsupervised graph partitioning method determines the communities of standards that are similar.
Moreover,  $\textit{I4.0}\cal{RD}$ explores communities to identify possible relations of standards, enhancing, thus, interoperability.

\section{Problem Definition and Proposed Solution}
\label{sec:approach}
\begin{figure*}[tb]
\centering
\vspace{0pt}
\subfloat[Actual I4.0 KG]
{\includegraphics[width=.45\linewidth]{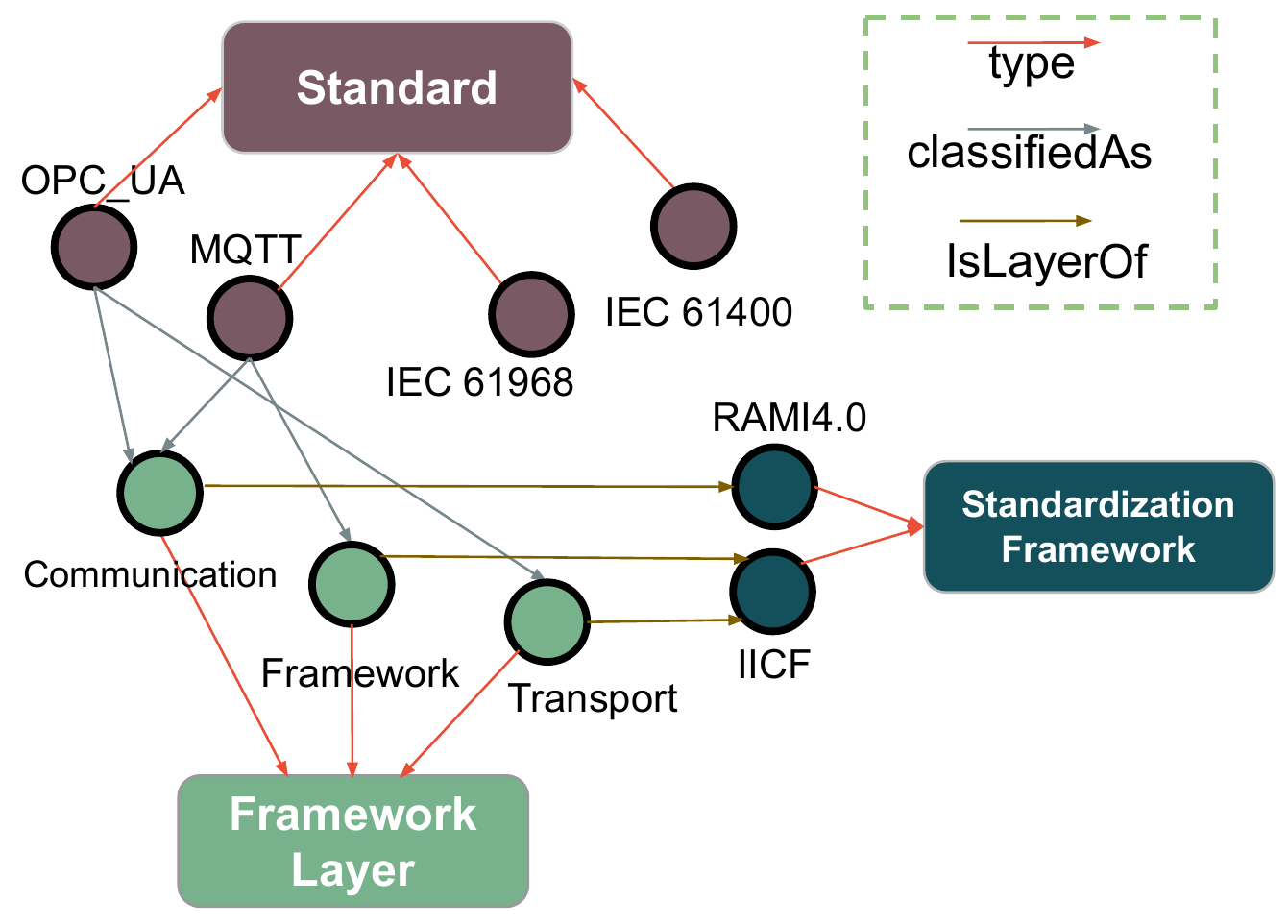}\label{fig:actual_kg}}
\vspace{0pt}
\subfloat[Ideal I4.0 KG]
{\includegraphics[width=.45\linewidth]{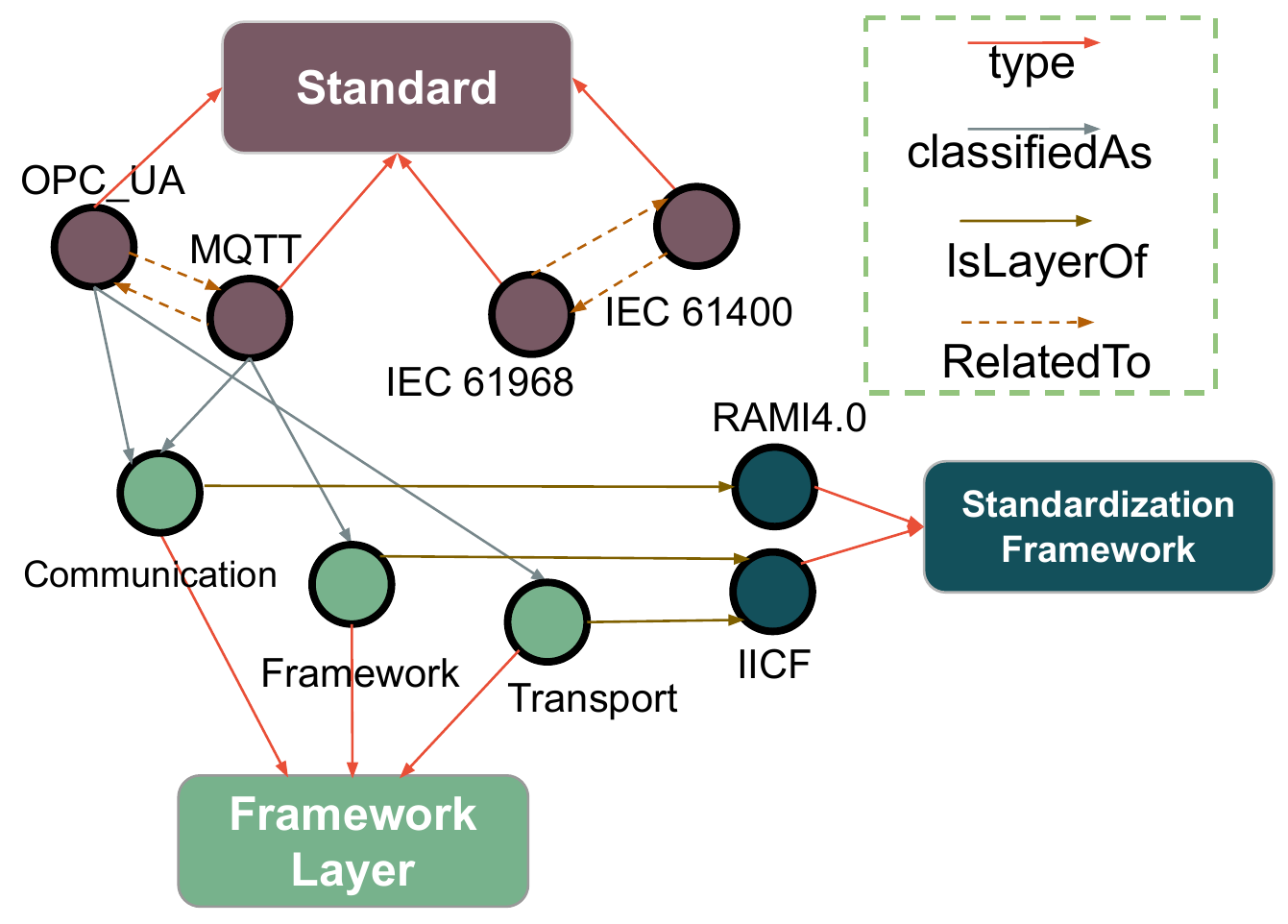}\label{fig:ideal_kg}} 
\caption{\textbf{Example of I4.0KGs}. Figure~\ref{fig:actual_kg} shows known relationships among standards to Framework Layer and Standardization Framework. While Figure~\ref{fig:ideal_kg} depicts all the ideal relationships between the standards expressed with the property \texttt{relatedTo}. Standards OPC UA and MQTT are related, as well as the standards IEC 61968 and IEC 61400. Our aim is discovering relations \texttt{relatedTo} in Figure~\ref{fig:ideal_kg}.}
\label{fig:sto}
\end{figure*}

We tackle the problem of unveiling relations between I4.0 standards.
We assume that the relations among standards and standardization frameworks like the ones shown in Figure \ref{fig:sto}(a), are represented in a knowledge graph named I4.0KG. Nodes in a I4.0KG correspond to standards and frameworks; edges represent relations among standards, as well as the standards grouped in a framework layer. An I4.0KG is defined as follows:

%\begin{definition}[I4.0 Standard Knowledge Graph]
%\label{definition1}
%Let $U$ be a set of RDF URI references and $L$ a set of RDF literals.
Given sets $V_{e}$ and $V_{t}$ of entities and types, respectively, a set $E$ of labelled edges representing relations, and  a set $L$ of labels. An I.40KG is defined as $\cal{G}$ $=(V_{e} \cup V_{t}, E, L)$:
\begin{itemize}
\item[$\bullet$] The types Standard, Frameworks, and Framework Layer belong to $V_{t}$.
\item[$\bullet$] I4.0 standards, frameworks, and layers are represented as instances of $V_{e}$.
\item[$\bullet$] The types of the entities in $V_{e}$ are represented as edges in $E$ that belong to $V_{e} \times V_{t}$.
\item[$\bullet$] Edges in $E$ that belong to  $V_{e} \times V_{e}$ represent relations between standards and their classifications into layers according to a framework. 
\item[$\bullet$] \textit{RelatedTo}, \textit{Type}, \textit{classifiedAs}, \textit{IsLayerOf} correspond to labels in $L$ that represent the relations between standards, their type, their classification into layers, and the layers of a framework, respectively.   
\end{itemize}
%\end{definition}

\iffalse
\begin{definition}[Standard Community]
A Standard Community $\cal{SC}$=$(V_{a},E_{a},P_{a})$ corresponds to a subgraph of $I4.0KG = (V_{e} \cup V_{t}, E,P)$, where
\begin{itemize}
    \item[$\bullet$] Nodes are standard entities of type \emph{Standard},  
    \[V_{a}=\{a \mid (a\; \textit{rdf:type} \; \textit{:Standard}) \in E\}\]
    \item[$\bullet$] Standards are related according to \texttt{relatedTo} property, $E_{a}=\{(a_i \textit{ :relatedTo }a_j)\}$
    
    %$E_{a}=\{(a_i \textit{ :relatedTo }a_j) \mid \exists c $ . 
    %$ a_i, a_j \in V_{a} \wedge ( a_i \textit{ :hasClassification } c ) \in E %\wedge ( a_j \textit{ :hasClassification } \;c ) \in E \wedge (c\; %\textit{ rdf:type } \textit{ :StandardClassification } ) \in E \}$
\end{itemize}
\end{definition}
\fi
\subsection{Problem Statement}
Let $\cal{G}'$ $=(V_{e} \cup V_{t}, E',P)$ and $\cal{G}$ $= (V_{e} \cup V_{t}, E,P)$ be two I4.0 knowledge graphs. $\cal{G}'$ is an \emph{ideal} knowledge graph that contains all the \emph{existing relations} between standard entities and frameworks in $V_{e}$, i.e., an oracle that knows whether two standard entities are related or not, and to which layer they should belong; Figure \ref{fig:sto} (b) illustrates a portion of an ideal I4.0KG, where the relations between standards are explicitly represented. 
$\cal{G}$ $= (V_{e} \cup V_{t}, E,P)$ is an \emph{actual} I4.0KG, which only contains a portion of the relations represented in $\cal{G}'$, i.e., $E \subseteq E'$; it represents those relations that are known and is not necessarily complete. Let $\Delta(E', E) = E'- E$ be the set of relations existing in the ideal knowledge graph $\cal{G}'$ that are not represented in $\cal{G}$. 
Let $\cal{G}_\text{comp}$=$(V_{e} \cup V_{t}, E_\text{comp},P)$
be a \emph{complete} knowledge graph, which includes a relation for each possible combination of elements in $V_{e}$ and labels in $L$, i.e., $E\subseteq E'\subseteq E_\text{comp}$. Given a relation $e \in \Delta(E_\text{comp}, E)$, the problem of discovering relations consists of determining whether $e \in E'$, i.e., if a relation represented by an edge  $r$=$(e_i \; p \; e_j)$ corresponds to an existing relation in the ideal knowledge graph  $\cal{G}'$.
Specifically, we focus on the problem of discovering \textit{relations} between standards in $\cal{G}$ $= (V_{e} \cup V_{t}, E,P)$.
%Thus, we are interested in finding the Standard Community $\cal{SC}$ $=(V_{a},E_{a},P_{a})$ composed of the maximal set of relationships or edges that belong to the ideal I4.0 Knowledge Graph, i.e., the set $E_{a}$ in $\cal{SC}$ that corresponds to a solution of the following optimization problem:
We are interested in finding the maximal set of relationships or edges $E_{a}$ that belong to the ideal I4.0KG, i.e., find a set $E_{a}$ that corresponds to a solution of the following optimization problem:

\[\operatorname*{argmax}_{\mathit{E_{a}}\subseteq \mathit{E_{comp}}}{|E_{a} \cap E'|}\]

Considering the knowledge graphs depicted in Figures \ref{fig:sto} (a) and (b), the problem addressed in this work corresponds to the identification of edges in the ideal knowledge graph that correspond to unknown relations between standards.

\subsection{Proposed Solution}
We propose a relation discovery method over I4.0KGs to identify unknown relations among standards. Our proposed method exploits relations represented in an I4.0KG to compute similarity values between the modeled standards. Further, an unsupervised graph partitioning method determine the parts of the I4.0KG or communities of standards that are similar. Then, the \textit{homophily} prediction principle is applied in a way that similar standards in a community are considered to be related.  

%Considering that not all standards have a layer associated and that a standard can be related to layers of different standardization frameworks, we propose to build communities of standards based on their relationships.
%In this approach, standards that are related to each other could be in the same community.
%This facilitates the resolution of interoperability problems and also it assigns a layer to the standards that do not have any associated layer.

\section{The $\textit{I4.0}\cal{RD}$ Architecture}
\label{sec:architecture}
\begin{figure*}[t!]
  \centering    
  \includegraphics[width=\linewidth]{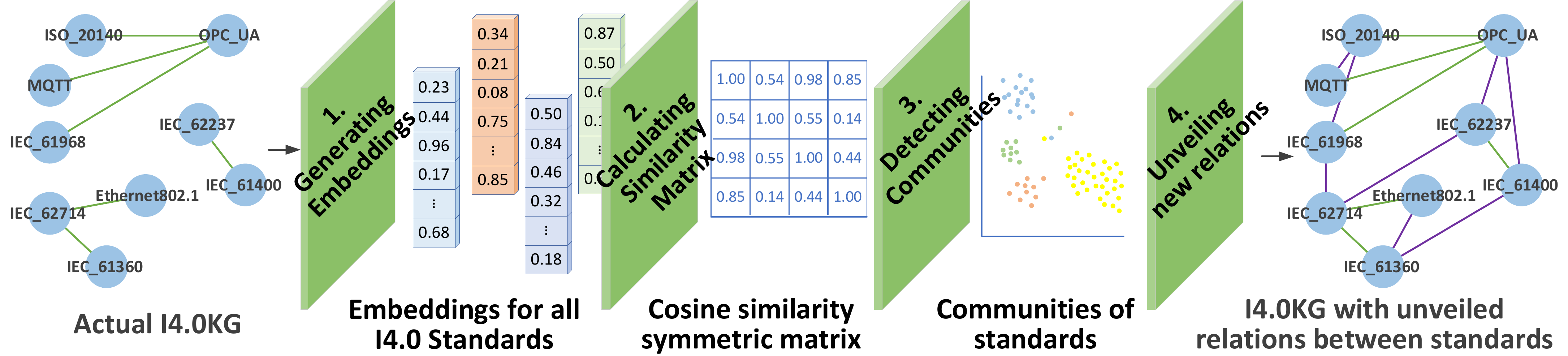}
  \caption{\textbf{Architecture.} $\textit{I4.0}\cal{RD}$ receives an I4.0KG and outputs an extended version of the I4.0KG including novel relations. Embeddings for each standard are created using the Trans* family of models, and similarity values between embeddings are computed; these values are used to partition standards into communities. Finally, the homophily prediction principle is applied to each community to discover unknown relations.} 
  \label{fig:architecture}
\end{figure*}

Figure~\ref{fig:architecture} presents $\textit{I4.0}\cal{RD}$, a pipeline that implements the proposed approach. $\textit{I4.0}\cal{RD}$ receives an I4.0KG $\cal{G}$, and returns an I4.0KG $\cal{G}'$ that corresponds to a solution of the problem of discovering relations between standards. 
First, in order to compute the values of similarity between the entities an I4.0KG, $\textit{I4.0}\cal{RD}$ learns a latent representation of the standards in a high-dimensional space. Our approach resorts to the Trans$^*$ family of models to compute the embeddings of the standards and the cosine similarity measure to compute the values of similarity. 
Next, community detection algorithms are applied to identify communities of related standards. METIS~\cite{Karypis:1998}, KMeans~\cite{Arthur:2007}, and SemEP~\cite{Palma:2014} are methods included in the pipeline to produce different communities of standards.
Finally, $\textit{I4.0}\cal{RD}$ applies the \textit{homophily} principle to each community to predict relations or alignments among standards.

%Finally, analytic algorithms are executed on the identified communities to evaluate the identified patterns and links; experts are kept in the loop.

\subsection{Learning Latent Representations of Standards}
\label{embeddings-creation}
$\textit{I4.0}\cal{RD}$ utilizes the  Trans$^*$ family of models to compute latent representations, e.g., vectors, of entities and relations in an I4.0 knowledge graph.
In particular, $\textit{I4.0}\cal{RD}$ utilizes TransE, TransD, TransH, and TransR.
%Explain the differences
These models differ on the representation of the embeddings for the entities and relations (Wang et al.~\cite{WangMWG17}).
%The variables h, r, t denote the head entity, relation, and the tail entity, respectively.
Suppose $e_i$, $e_j$, and $p$, denote the vectorial representation of two entities related by the labeled edge $p$ in an I4.0 knowledge graph. Furthermore, $\|x\|_{2}$ represents the Euclidean norm.

TransE, TransH, and TranR represent the entity embeddings as $(e_i,e_j \in \mathbb{R}^d)$, while TransD characterizes the entity embeddings as: $(e_i,w_{e_i} \in \mathbb{R}^d - e_i,w_{e_j} \in \mathbb{R}^d )$.
As a consequence of different embedding representations, the scoring function also varies. For example, TransE is defined in terms of the score function $\|e_i + p - e_j\|_2^2$ , while $\|M_p e_i + p - M_p e_j\|_2^2$ defines TransR\footnote{$M_p$ corresponds to a projection matrix $M_p \in \mathbb{R}^{dxk}$ that projects entities from the entity space to the relation space; further $p \in \mathbb{R}^k$.}.
Furthermore, TransH score function corresponds to $\|{e_i}_\perp + d_p - {e_j}_\perp\|_2^2$, where the variables  ${e_i}_\perp$ and ${e_j}_\perp$ denote a projection to the hyperplane $w_p$ of the labeled relation p, and
$d_p$ is the vector of a relation-specific translation in the hyperplane $w_p$.
To learn the embeddings, $\textit{I4.0}\cal{RD}$ resorts to the PyKeen (Python KnowlEdge EmbeddiNgs) framework~\cite{Ali2019}.
As hyperparameters for the models of the Trans$^*$ family, we use the ones specified in the original papers of the models.
The hyperparameters include embedding dimension (set to 50), number of epochs (set to 500), batch size (set to 64), seed (set to 0), learning rate (set to 0.01), scoring function (set to 1 for TransE, and 2 for the rest), margin loss (set to 1 for TransE and 0.05 for the rest).
%Listing~\ref{lst:code} shows an excerpt from the Python class that configures the TransD model in PyKeen.
All the configuration classes and hyperparameters are open in GitHub \furl{https://github.com/i40-Tools/I4.0KG-Embeddings}.

\begin{comment}

\begin{lstlisting}[language=python, escapechar=@, caption={\textbf{Code to configure TransD model in PyKeen}}, label={lst:code}]
from pykeen.kge_models import TransD
 
output_directory = './embeddings/TransD/I4.0KG'

config = dict(
    training_set_path           = './I4.0KG/I4.0KG.nt',
    execution_mode              = 'Training_mode',
    random_seed                 = 0,
    kg_embedding_model_name     = 'TransD',
    embedding_dim               = 50,
    relation_embedding_dim      = 20,
    scoring_function            = 2,  #corresponds to L2
    margin_loss                 = 0.05,
    learning_rate               = 0.01,
    num_epochs                  = 100,
    batch_size                  = 64,
    preferred_device            = 'gpu'
)
results = pykeen.run(
    config=config,
    output_directory=output_directory,
)
\end{lstlisting}
\end{comment}

\iffalse
\begin{figure*}[t!]
  \centering    
  \includegraphics[width=\linewidth]{images/Measure_standards_clusters_based_on_Trans_Family.pdf}
  \caption{Quality of the generated communities.  Communities evaluated in terms of prediction metrics with thresholds of 0.85, 0.90, and 0.95 using the SemEP, METIS, and KMeans algorithms. In this case higher values are better. Our approach exhibits the best performance with value 0.85 for the threshold and using TransE to create the embeddings.} 
  \label{fig:quality_communities_relatedTo2}
\end{figure*}
\fi

\subsection{Computing Similarity Values Between Standards}
\label{methods}
Once the algorithm--Trans$^*$ family--that computes the embeddings reaches a termination condition, e.g., the maximum number of epochs, the I4.0KG embeddings are learned. 
As the next step, $\textit{I4.0}\cal{RD}$ calculates a \emph{similarity symmetric matrix} between the embeddings that represent the I4.0 standards.
Any distance metric for vector spaces can be utilized to calculate this value. However, as a proof of concepts, $\textit{I4.0}\cal{RD}$ applies the Cosine Distance.
Let $u$ be an embedding of the Standard-A and $v$ an embedding of the Standard-B, the similarity score, between both standards, is defined as follows:
       \[
    cosine(u,v) = 1 - \dfrac{u . v}{||u||_2 ||v||_2}
    \]
After building the \emph{similarity symmetric matrix}, $\textit{I4.0}\cal{RD}$ applies a threshold to restrict the similarity values.
$\textit{I4.0}\cal{RD}$ relies on percentiles to calculate the value of such a threshold.
Further, $\textit{I4.0}\cal{RD}$ utilizes the function Kernel Density Estimation (KDE) to compute the probability density of the cosine similarity matrix; it sets to zero the similarity values lower than the given threshold.

\subsection{Detecting Communities of Standards}
$\textit{I4.0}\cal{RD}$ maps the problem of computing groups of potentially related standards to the problem of community detection. 
Once the embeddings are learned, the standards are represented in a vectorial way according to their functions preserving their semantic characteristics. 
Using the embeddings,  $\textit{I4.0}\cal{RD}$ computes the similarity between the I4.0 standards as mentioned in the previous section.
The values of similarity between standards are utilized to partition the set of standards in a way that standards in a community are highly similar but dissimilar to the standards in other communities.
As proof of concept, three state-of-the-art community detection algorithms have been used in $\textit{I4.0}\cal{RD}$: SemEP, METIS, and KMeans. They implement diverse strategies for partitioning a set based on the values of similarity, and our goal is to evaluate which of the three is more suitable to identify meaningful connections between standards. 

\subsection{Discovering Relations Between Standards}
\label{discovery}
\begin{figure*}[h]
\centering
\vspace{0pt}
\subfloat[Application of the Homophily Prediction Principle]
{\includegraphics[width=.45\linewidth]{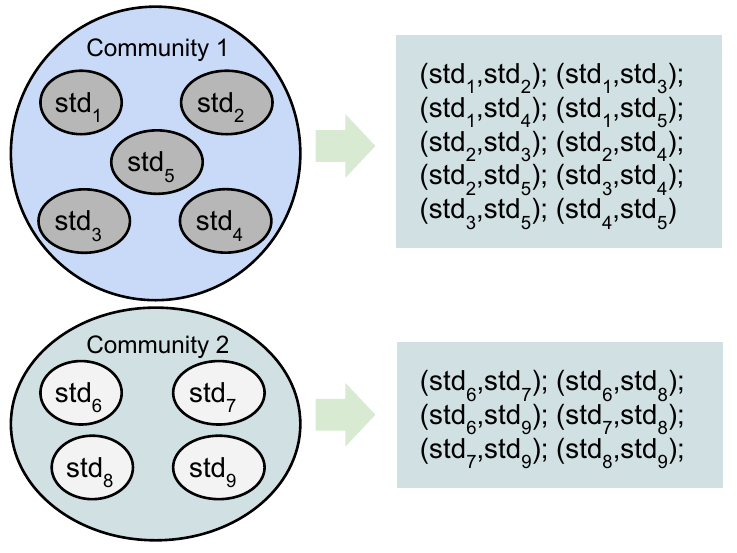}\label{fig:communities}} 
\vspace{0pt}
\subfloat[Known Relations used to determine discovered relations between standards]
{\includegraphics[width=.30\linewidth]{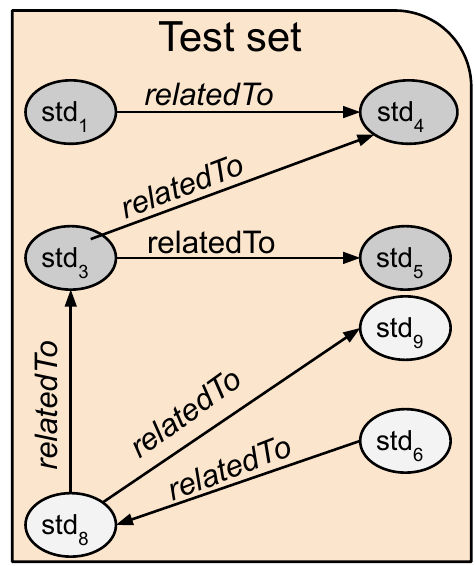}\label{fig:testset}} 
\vspace{0pt}
\caption{\textbf{Discovering Relations Between Standards}.
(a) The homophily prediction principle is applied on two communities, as a result, 16 relations between standards are found. (b) Six out of the 16 found relations correspond to meaningfully relatisons.}
\label{fig:exampleC}
\end{figure*}

New relations between standards are discovered in this step; 
the \emph{homophily} prediction principle is applied over each of the communities and all the standards in a community are assumed to be related. Figure~\ref{fig:exampleC} depicts an example where new relations are computed from two communities; unknown relations correspond to connections between standards in a community that did not existing in the input I4.0KG.

\section{Empirical Evaluation}
\label{sec:evaluation}
We report on the impact that the knowledge encoded in I4.0 knowledge graphs has in the behavior of $\textit{I4.0}\cal{RD}$.
In particular, we asses the following research questions:
\begin{enumerate}
    \item[\textbf{RQ1)}] Can the semantics encoded in I4.0KG empower the accuracy of the relatedness between entities in a KG?
    \item[\textbf{RQ2)}] Does a semantic community based analysis on I4.0KG allow for improving the quality of predicting new relations on the I4.0 standards landscape?
\end{enumerate}

\textbf{Experiment Setup:}
We considered four embedding algorithms to build the standards embedding. Each of these algorithms was evaluated independently.
Next, a similarity matrix for the standards embedding was computed. The similarity matrix is required for applying the community detection algorithms. In our experiments, three algorithms were used to compute the communities.
That means twelve combinations between embedding algorithms and community detection algorithms to be evaluated.
To assure statistical robustness, we executed 5-folds cross-validation with one run.

\begin{figure*}[tb]
\centering
\vspace{0pt}
\subfloat[\textbf{TransD-Density in 5-fold}]
{\includegraphics[width=.35\linewidth]{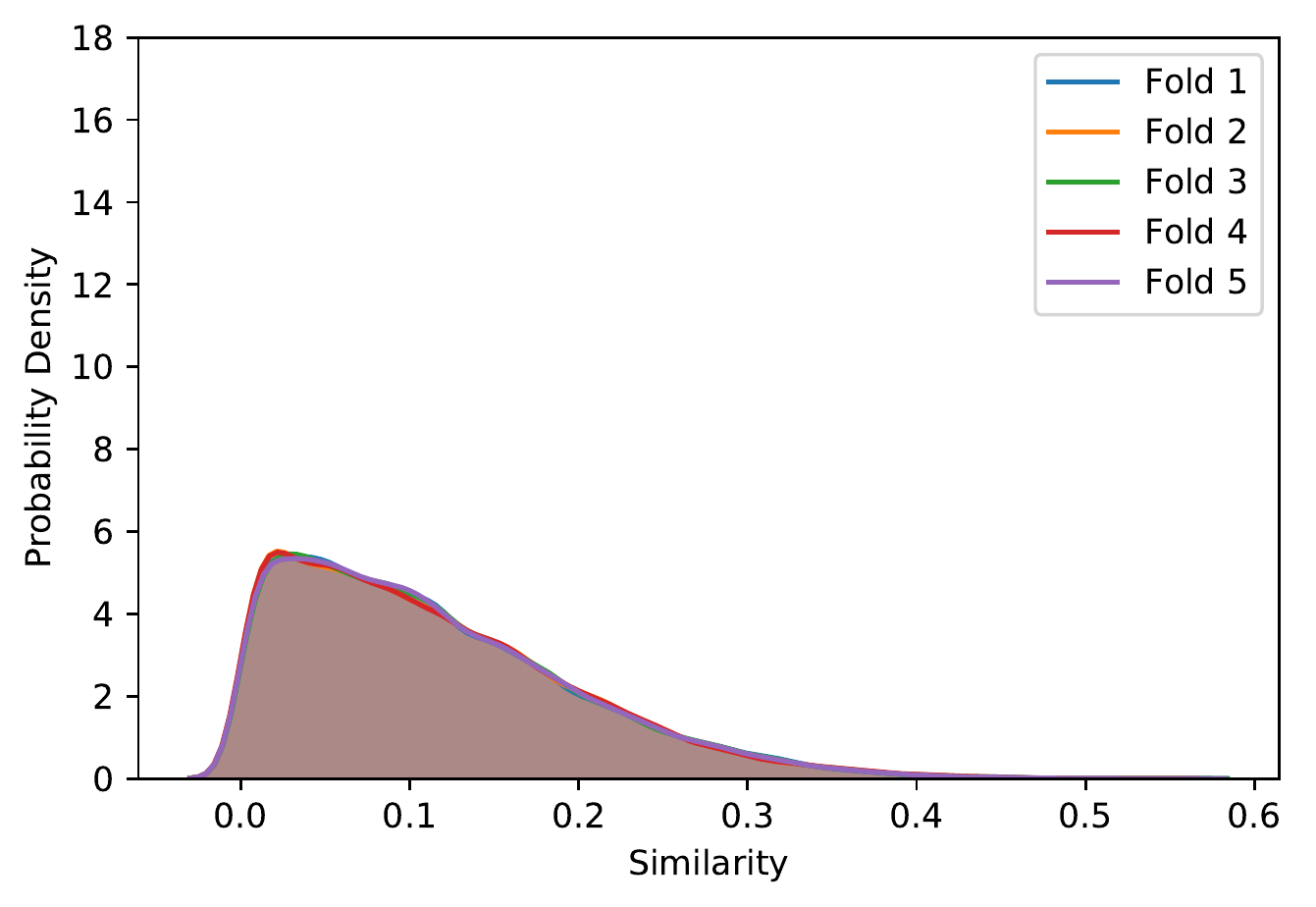}\label{fig:TrnasD_density}}
\vspace{0pt}
\subfloat[\textbf{TransE-Density in 5-fold}]
{\includegraphics[width=.35\linewidth]{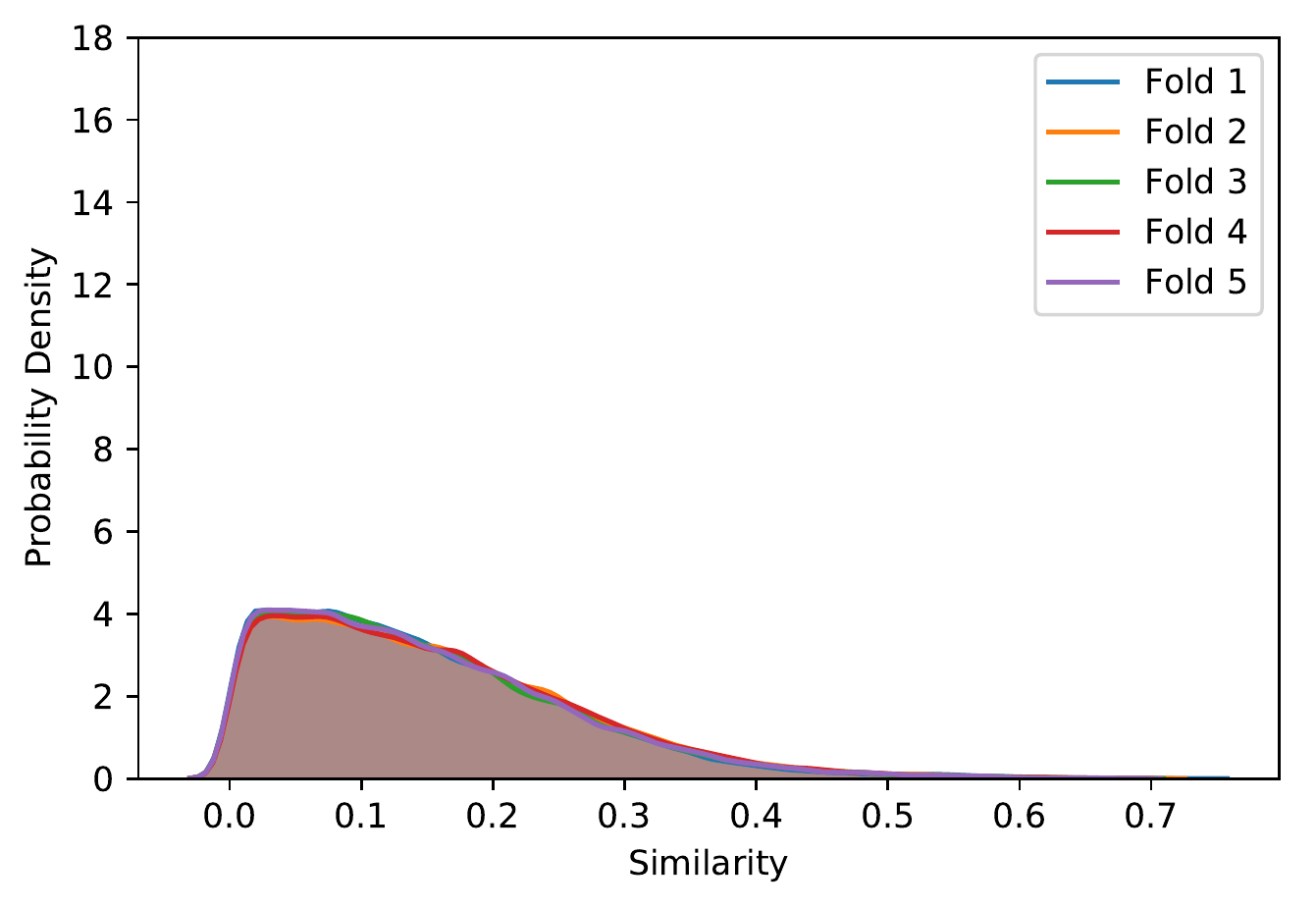}\label{fig:TrnasE_density}} 
\vspace{0pt}
\subfloat[\textbf{TransH-Density in 5-fold}]
{\includegraphics[width=.35\linewidth]{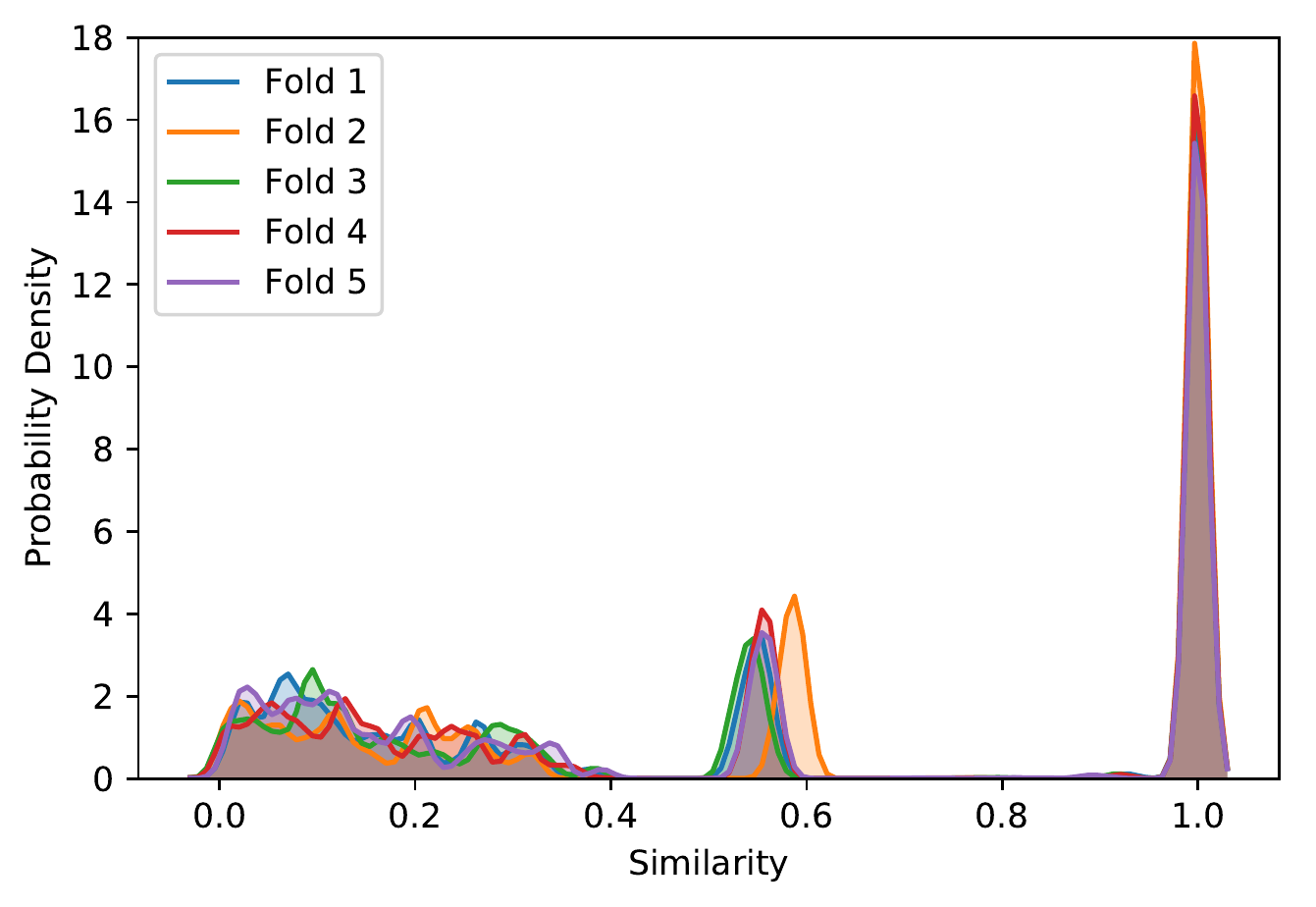}\label{fig:TrnasH_density}} 
\vspace{0pt}
\subfloat[\textbf{TransR-Density in 5-fold}]
{\includegraphics[width=.35\linewidth]{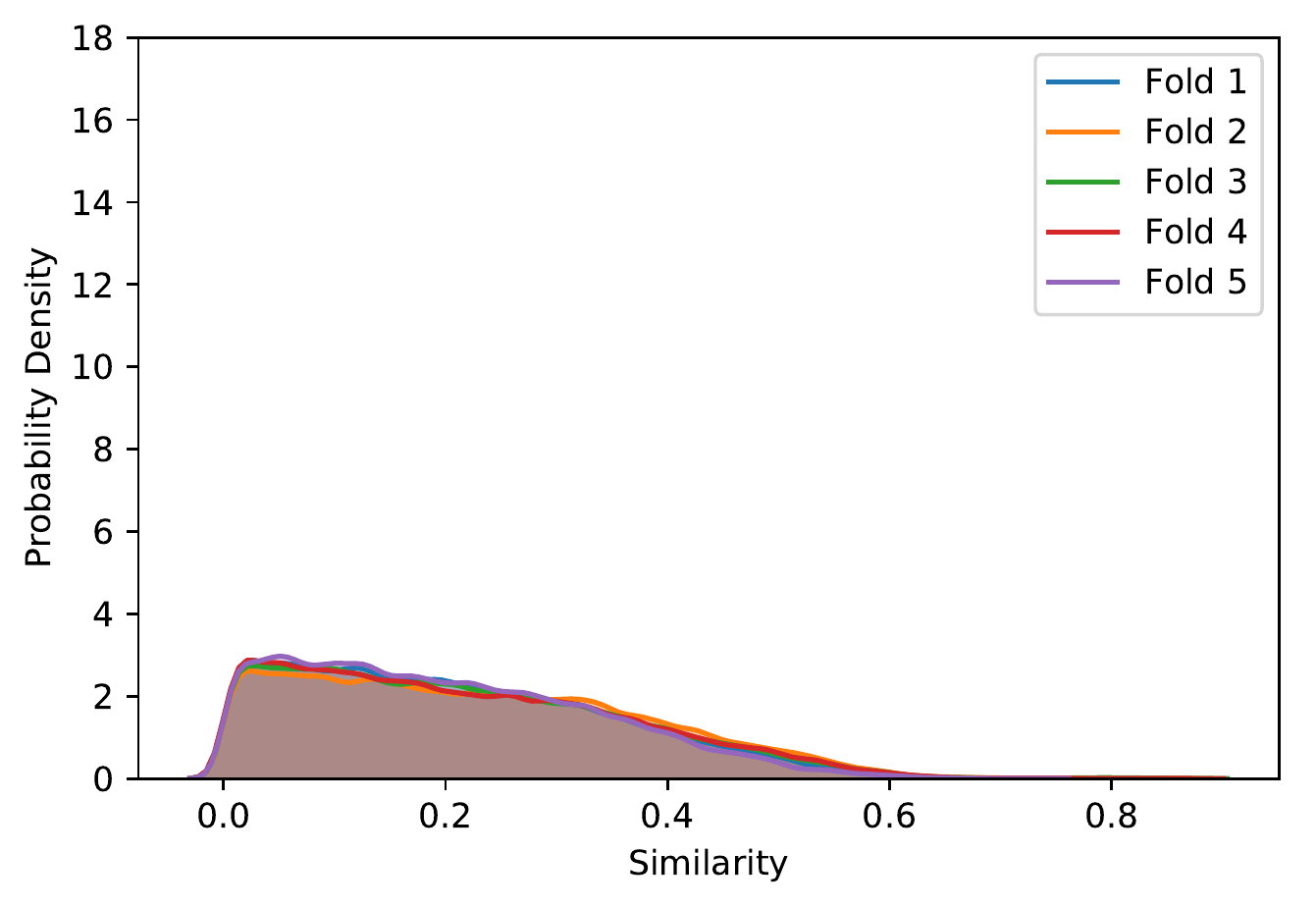}\label{fig:TrnasR_density}}
\caption{\textbf{Probability density of each fold per Trans$^*$ methods}. 
%Probability density for each five-fold and embedding algorithm. 
Figures~\ref{fig:TrnasD_density},~\ref{fig:TrnasE_density},and ~\ref{fig:TrnasR_density} show that all folds have values close to zero, i.e., with embeddings created by TransD, TransE, and TransR the standards are very different from each other.
However, TransH (cf. Figure~\ref{fig:TrnasH_density}), exploits properties of the standards and generates embeddings with a different distribution of similarity, i.e., values between 0.0 and 0.6, as well as values close to 1.0.
According to known characteristics of the I4 standards, the TransH distribution of similarity better represents their relatedness.}
\label{fig:density_standards_similarity}
\end{figure*}

\textbf{Thresholds for Computing Values of Similarity}
Figure~\ref{fig:density_standards_similarity} depicts the probability density function of each fold for each embedding algorithm.
Figures~\ref{fig:TrnasD_density} and \ref{fig:TrnasE_density} show the values of the folds of TransD and TransE where all the similarity values are close to 0.0, i.e., all the standards are different. 
Figure~\ref{fig:TrnasR_density} suggests that all the folds have similar behavior with values between 0.0 and 0.5. 
Figure~\ref{fig:TrnasH_density} shows a group of standards similar with values close to 1.0 and the rest of the standards between 0.0 and 0.6.
The percentile of the similarity matrix is computed with a threshold of $0.85$.
That means all values of the similarity matrix which are less than the percentile computed, are filled with 0.0 and then, these two standards are dissimilar.
After analyzing the probability density of each fold (cf. Figure~\ref{fig:density_standards_similarity}), the thresholds of TransH and TransR are set to $0.50$ and $0.75$, respectively.
The reason is because the two cases with a high threshold find all similar standards. 
In the case of TransH, there is a high density of values close to 1.0; it indicates that for a threshold of 0.85, the percentile computed is almost 1.0. the values of the similarity matrix less than the threshold are filled with 0.0; values of 0.0 represent that the compared standards are not similar.

\textbf{Metrics:} the following metrics are used to estimate the quality of the communities from the I4.0KG embeddings. 
\begin{itemize}
    \item[a)] \textbf{Conductance (InvC)}: measures relatedness of entities in a community, and how different they are to entities outside the community~\cite{Gaertler:2005}. 
    The inverse of Conductance is reported: $1 - Conductance(K)$, where $K = \{k_1, k_2, ...., k_n\}$ the set of standards communities obtained by the cluster algorithm, and $k_i$ are the computed clusters.
    \item[b)] \textbf{Performance (P)}: sums up the number of intra-community relationships, plus the number of non-existent relationships between communities~\cite{Gaertler:2005}. 
    \item[c)] \textbf{Total Cut (InvTC)}: sums up all similarities among entities in different communities~\cite{inbook}.
The Total Cut values are normalized by dividing the sum of the similarities between the entities. The inverse of Total Cut is reported as follows: $1 - NormTotalCut(K)$ 
    \item[d)] \textbf{Modularity (M)}: is the value of the intra-community similarities between the entities divided by the sum of all the similarities between the entities, minus the sum of the similarities among the entities in different communities, in case they are randomly distributed in the communities~\cite{Newman8577}. 
The value of the Modularity is in the range of $[-0.5, 1]$, which can be scaled to $[0, 1]$ by computing: \( \frac{Modularity(K) + 0.5}{1.5}\). 
 \item[e)] \textbf{Coverage (Co)}: compares the fraction of intra-community similarities between entities to the sum of all similarities between entities~\cite{Gaertler:2005}.
\end{itemize}

\begin{figure*}[tb]
\centering
\vspace{0pt}
\subfloat[TransD - th:85]
{\includegraphics[width=.35\linewidth]{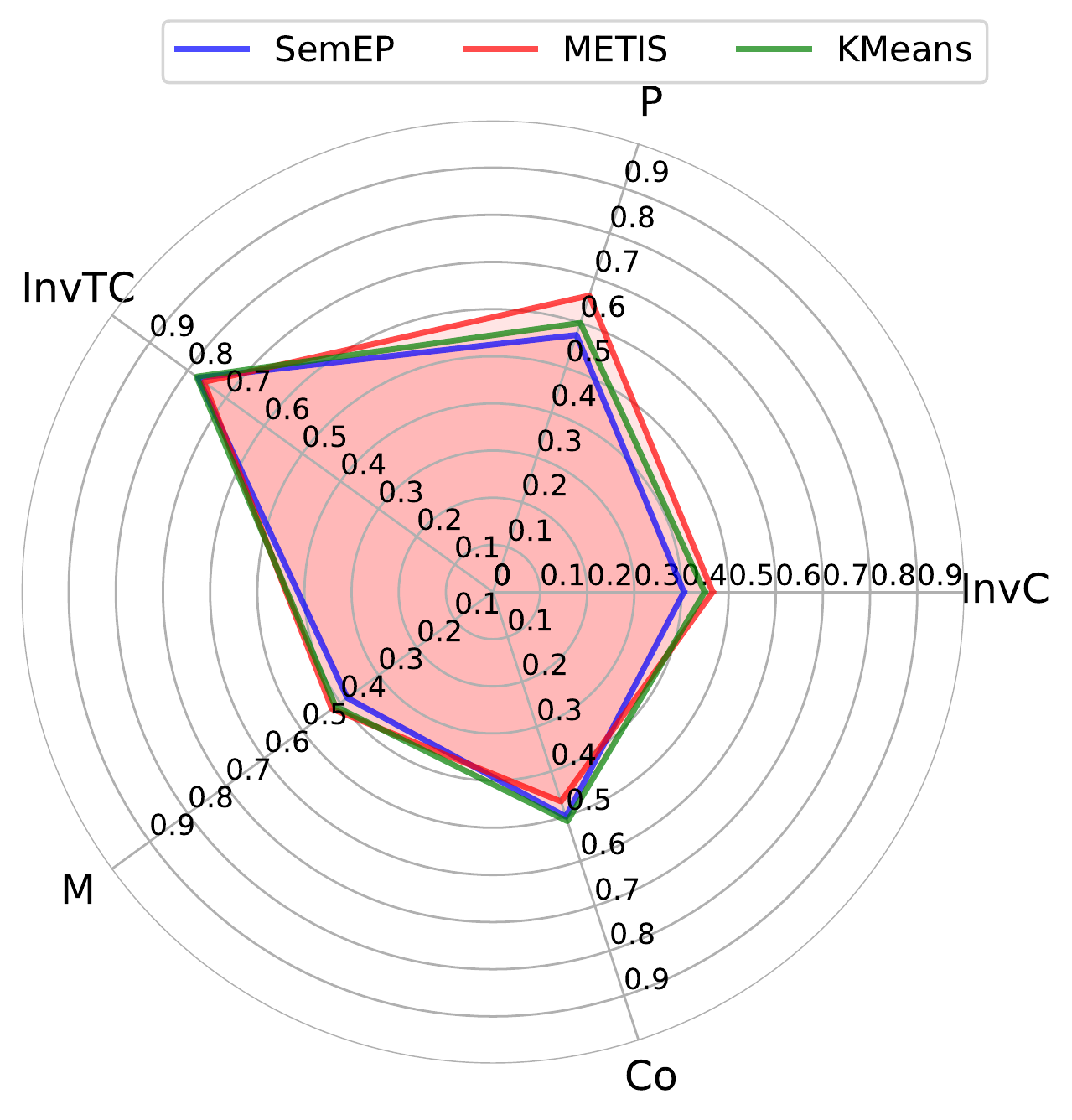}\label{fig:trnasD_85}}
\vspace{0pt}
\subfloat[TransE - th:85]
{\includegraphics[width=.35\linewidth]{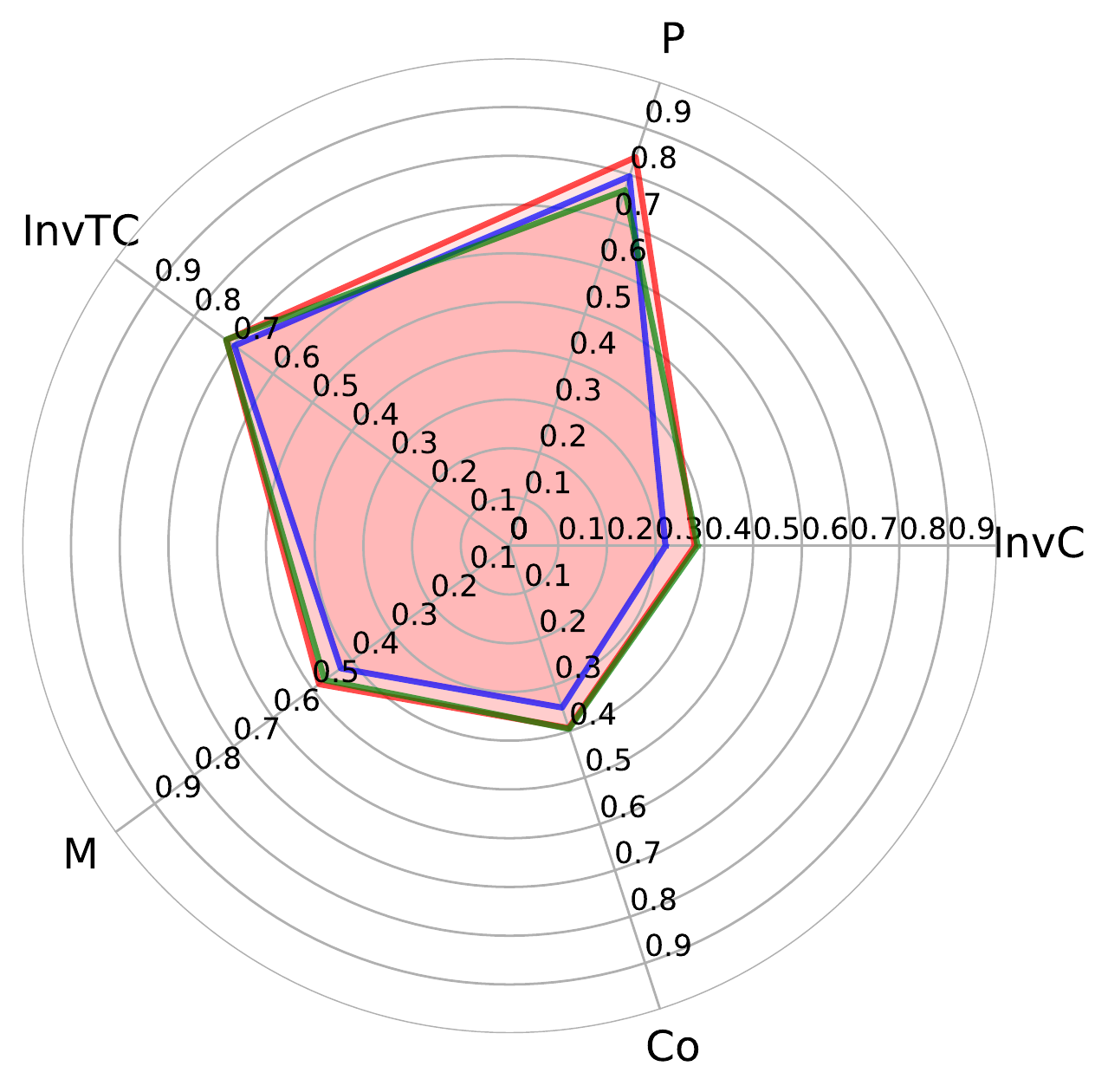}\label{fig:trnasE_85}} 
\\
\vspace{0pt}
\subfloat[TransH - th:50]
{\includegraphics[width=.35\linewidth]{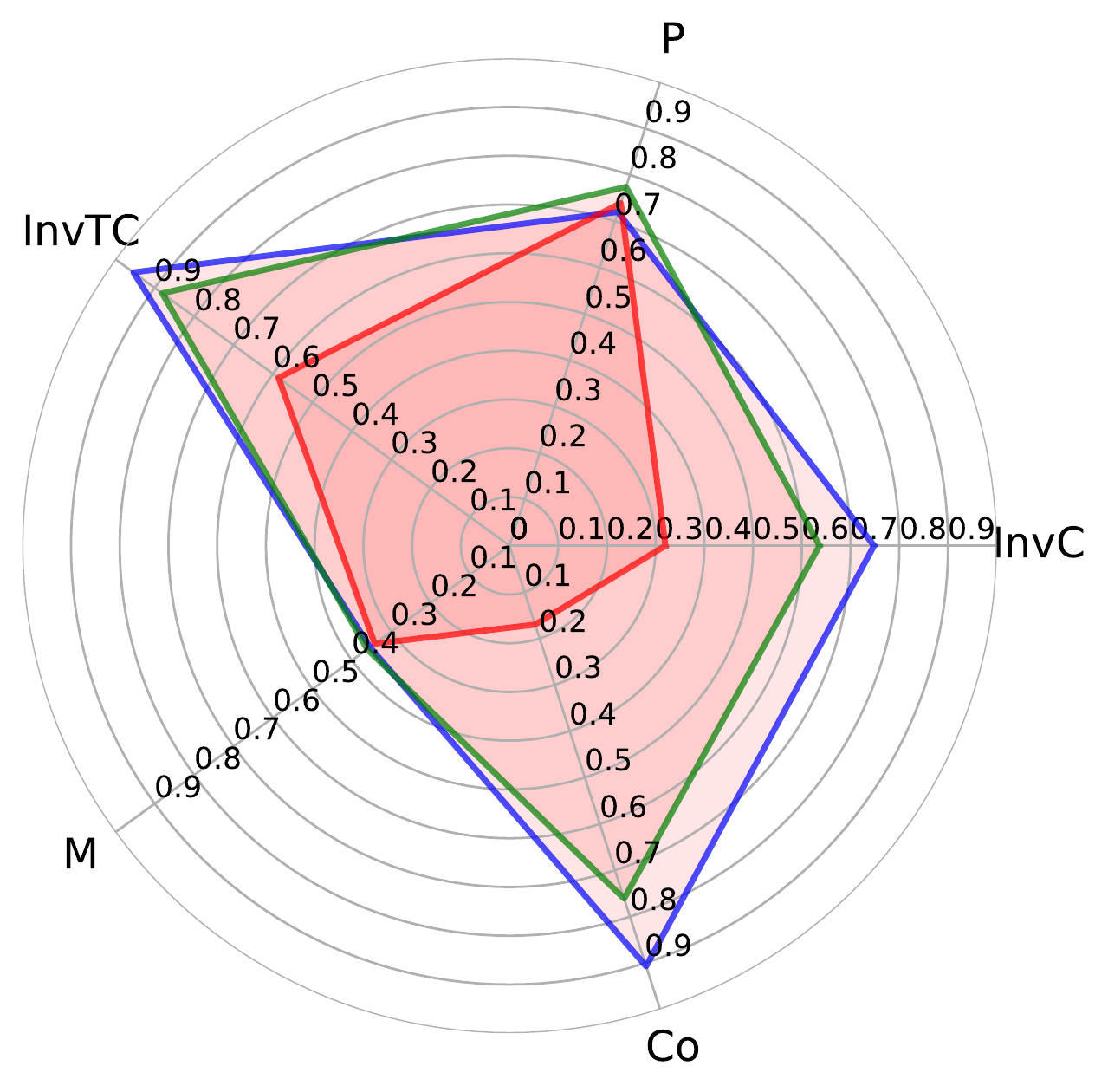}\label{fig:trnasH_50}} 
\vspace{0pt}
\subfloat[TransR - th:75]
{\includegraphics[width=.35\linewidth]{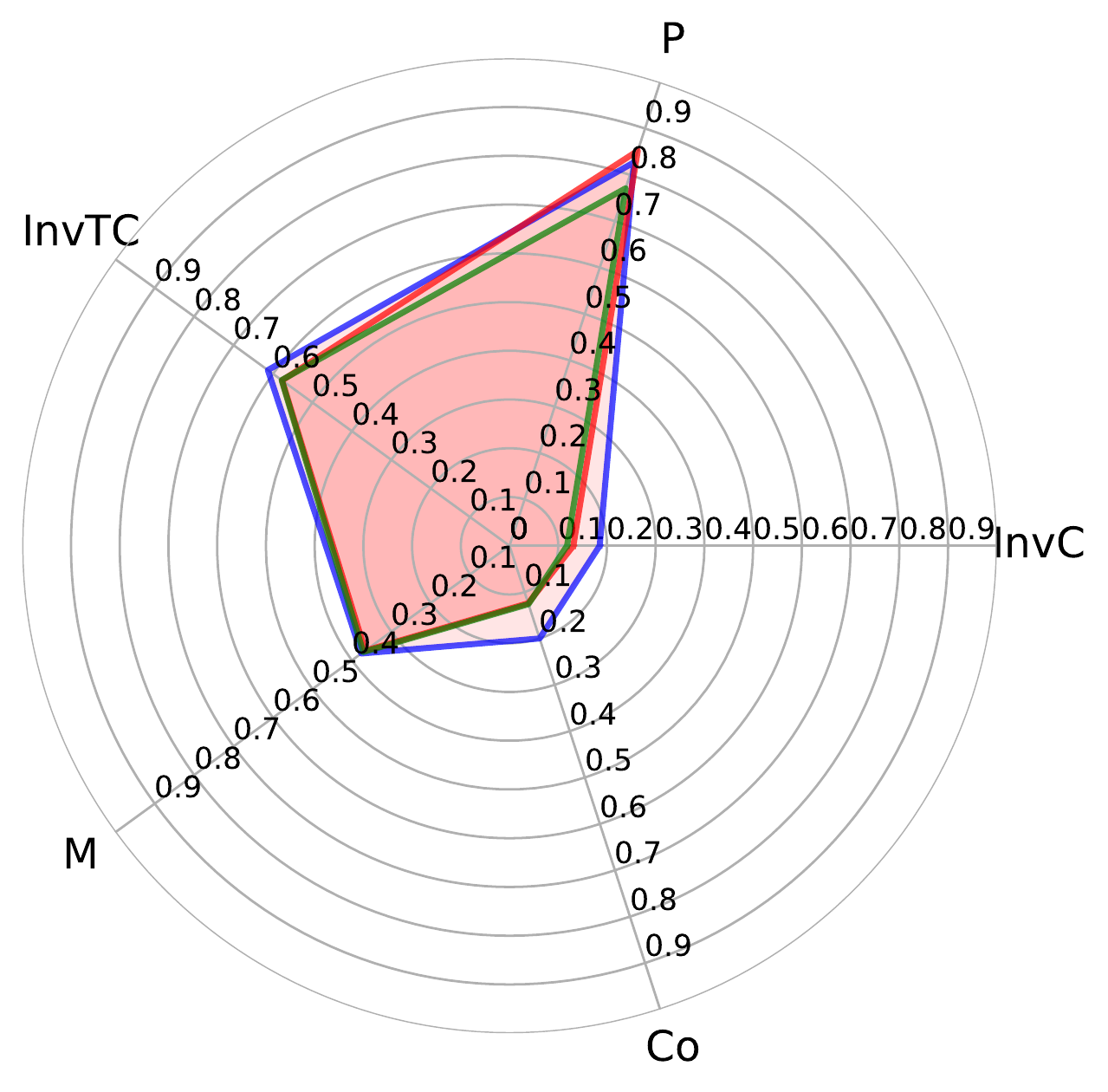}\label{fig:trnasR_75}}
\caption{\textbf{Quality of the generated communities}.  Communities evaluated in terms of prediction metrics with thresholds (th) of 0.85, 0.50, and 0.75 using the SemEP, METIS, and KMeans algorithms. 
In this case higher values are better. 
Our approach exhibits the best performance with TransH embeddings and a threshold of 0.50 for computing the similarity matrix, i.e., Figure (c). 
SemEP achieves the highest values in four of the five evaluated parameters.}
\label{fig:quality_communities_relatedTo}
\end{figure*}
\paragraph{\bf Implementation:} Our proposed approach is implemented in Python 2.7 and integrated with 
the PyKeen (Python KnowlEdge EmbeddiNgs) framework~\cite{Ali2019}, METIS 5.1~\footnote{\url{http://glaros.dtc.umn.edu/gkhome/metis/metis/download}},
SemEP~\footnote{\url{https://github.com/SDM-TIB/semEP}}, and Kmeans~\footnote{\url{https://scikit-learn.org/stable/modules/generated/sklearn.cluster.KMeans.html}}. 
The experiments were executed on a GPU server with ten chips Intel(R) Xeon(R) CPU E5-2660, two chips GeForce GTX 108, and 100 GB RAM.  
%The results of the experiments are published at~\furl{https://github.com/i40-Tools/I40KG-Embeddings}.
\paragraph{\textbf{RQ1} - Corroborating the accuracy of relatedness between standards in I40KG.}
To compute accuracy of $\textit{I4.0}\cal{RD}$, we executed a five-folds cross-validation procedure. 
To that end, the data set is divided into five consecutive folds shuffling the data before splitting into folds.
Each fold is used once as validation, i.e., test set while the remaining fourth folds form the training set.
Figure~\ref{fig:quality_communities_relatedTo} depicts the best results are obtained with the combination of the TransH and SemEP algorithms. 
The values obtained for this combination are as follows: \textbf{Inv. Conductance} ($0.75$), \textbf{Performance} ($0.77$), \textbf{Inv. Total Cut} ($0.95$), \textbf{Modularity} ($0.36$), and \textbf{Coverage} ($0.91$).

\begin{figure}[h]
\centering
%\vspace{0pt}
{\includegraphics[width=0.85\linewidth]{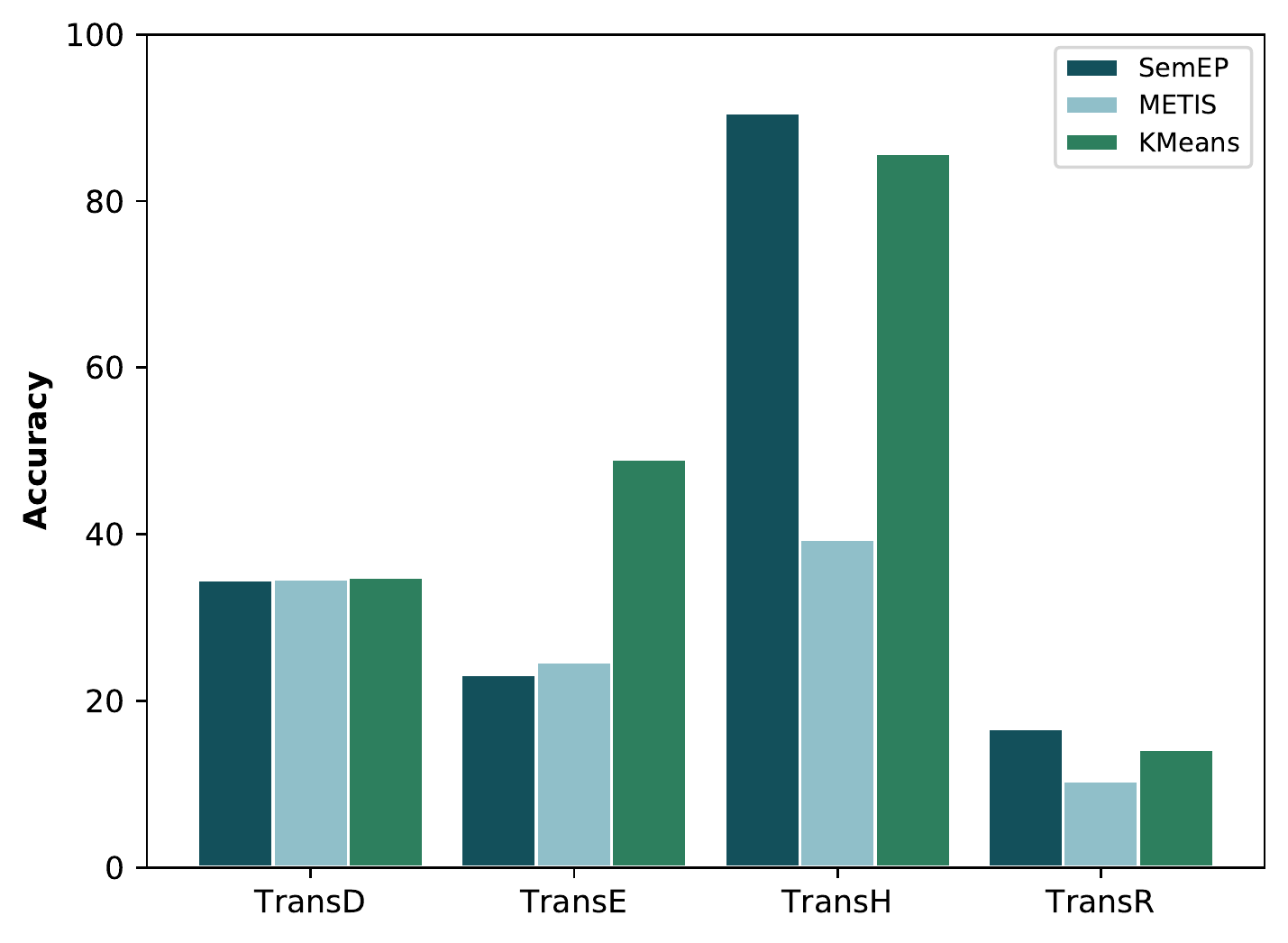}}
\caption{\textbf{$\textit{I4.0}\cal{RD}$ accuracy}. Percentage of the test set for the property \texttt{relatedTo} is achieved in each cluster. 
Our approach exhibits the best performance using TransH embedding and with the SemEP algorithm reaching an accuracy by up to 90\%.}
\vspace{-0.5 cm}
\label{fig:accuracy_of_related_standards}
\end{figure}

\paragraph{\textbf{RQ2} - Predicting new relations between standards.}
In order to assess the second research question, the data set is divided into five consecutive folds.
Each fold comprises 20\% of the relationships between standards. 
Next, the precision measurement is applied to evaluate the main objective is to unveil uncovered associations and at the same time to corroborate knowledge patterns that are already known.
%In order to assess the second research question, a test set of the I4.0KG is created.
%The test set comprises the 30\% of the relationships between standards. 
%This test set is built of a list of unique elements chosen from the population. 
%For this purpose, we apply random sampling without replacement.
%Then, the triples belonging to the 30\% are removed from the KG. 
\iffalse
\begin{algorithm}
	\caption{Compute accuracy of communities based on hasClassification property}
	\label{alg:accuracy_hasClassification}
	\begin{flushleft}
         \textbf{Input:} communities, test-set (hasClassification) \\
	    \textbf{Output:} accuracy (acc)
    \end{flushleft}
	\begin{algorithmic}[1]
		\Procedure{Compute accuracy of the cluster}{}
		
		\For{{$cluster$ in $communities$}} :
		\State $Select$ the standards of the test-set that are in $cluster$. $S = f($test-set$) \cup f(cluster)$ 
		\State $Select$ the classification of $S$ from test-set. $P$
		\State $Select$ the classification of $cluster$ from $cluster$. $C$
		\State $Interception$ of $P$ and $C$ classification. $I = f(P) \cup f(C)$
		\State $Normalize$ the result of $I$ to the scale of $1$. $acc = \frac{I}{f(P)}$
		\EndFor
		\State Return $acc$
		\EndProcedure
	\end{algorithmic}
\end{algorithm}
\fi
As shown in Figure~\ref{fig:accuracy_of_related_standards}, the best results for the property \texttt{relatedTo} are achieved by TransH embeddings in combination with the SemEP and KMeans algorithm.

The communities of standards discovered using the techniques TransH and SemEP contribute to the resolution of interoperability in I4.0 standards. 
To provide an example of this, we observed a resulting cluster with the standards \emph{ISO 15531} and \emph{MTConnect}. 
The former provides an information model for describing manufacturing data. 
The latter offers a vocabulary for manufacturing equipment. 
It is important to note that those standards are not related to the training set nor in I40KG. 
The membership of both standards in the cluster means that those two standards should be classified together in the standardization frameworks. 
Besides, it also suggests to the creators of the standards that they might look after possible existing synergies between them.
This example suggests that the techniques employed in this work are capable of discovering new communities of standards. 
These communities can be used to improve the classification that the standardization frameworks provide for the standards.
  
\iffalse
\begin{table}[]
    \centering
    \begin{tabular}{c}
\hline 
\textbf{Standards } \tabularnewline
\hline 
$IEC\_60839\_P7\_S1\_E1$ \tabularnewline
$IEC\_60839\_P7\_S2\_E1$ \tabularnewline
$IEC\_60839\_P7\_S3\_E1$ \tabularnewline
$IEC\_60839\_P7\_S4\_E1$ \tabularnewline
$IEC\_60839\_P7\_S5\_E1$ \tabularnewline
$IEC\_60839\_P7\_S6\_E1$  \tabularnewline
$IEC\_60839\_P7\_S7\_E1$  \tabularnewline
\hline 
\end{tabular}
    \caption{\textbf{Standards of alarm transmission systems}. These standards have been classified by our approach in the same community. The discovered communities can help the standardization frameworks to a better classification of the standards based on their features, thus reducing interoperability in the I4.0 context.}
    \label{tab:sto_alarm_transmission_systems}
\end{table}
\fi

\subsection{Discussion}
\label{discussion}
% the I40KG we use the property transitiva y simetrica
% Esto hace que se generen relaciones muchos a muchos
% TransE has limitations with these type of relations. 
% Therefore, TransE 
The techniques proposed in this paper rely on known relations between I4.0 standards to discover novel patterns and new relations.
During the experimental study, we can observe that these techniques could group together not only standards that were known to be related, but also standards whose relatedness was implicitly represented in the I40KG. 
This feature facilitates the detection of high-quality communities as reported in Figure \ref{fig:quality_communities_relatedTo}, as well as for an accurate discovery of relations between standards (cf. Figure \ref{fig:accuracy_of_related_standards}). 
As observed, the accuracy of the approach can be benefited from the application of state-of-the-art algorithms of the Trans$^*$ family, e.g., TransH. 
Additionally, the strategy employed by SemEP that allows for positioning in the same communities highly similar standards, leads our approach into high-quality discoveries.
The combination of both techniques TransH and SemEP allows discovering communities with high quality.

To understand why the combination of TransH and SemEP produces the best results, we analyze in detail both techniques.
TransH introduces the mechanism of projecting the relation to a specific hyperplane~\cite{Wang2014KnowledgeGE}, enabling, thus, the representation of relations with cardinality many to many. Since the materialization of transitivity and symmetry of the property \texttt{relatedTo} corresponds to many to many relations, the instances of this materialization are taken into account during the generation of the embeddings, specifically, during the translating operation on a hyperplane. 
%TransH enables an standard to be represented in different vectors given the projection on the hyperplane generated %by the relation \texttt{sto:relatedTo}.
%Therefore, having one-to-one/many-to-one/one-to-many/many-to-many relations in the KG makes the TransH work with better accuracy than the remaining of the Trans* family.
%Thus, TransH outlines a better representation of these relations. 
Thus, even thought semantics is not explicitly utilized during the computation of the embeddings, considering different types of relations, empowers the embeddings generated by TransH. Moreover, it allows for a more precise encoding of the standards represented in I4.0KG. Figure~\ref{fig:TrnasH_density} illustrates groups of standards in the similarity intervals $[0.9, 1.0], [0.5, 0.6]$, and $[0.0, 0.4]$. 
The SemEP algorithm can detect these similarities and represent them in high-precision communities.
%Here add two sentences why the strategy employed by SemEP works better
The other three models embeddings TransD, TransE, and TransR do not represent the standards in the best way. Figures \ref{fig:TrnasD_density}, \ref{fig:TrnasE_density}, \ref{fig:TrnasR_density} report that several standards are in the similarity interval $[0.0, 0.3]$. This means that no community detection algorithm could be able to discover communities with high quality.
Reported results indicate that the presented approach enables -- in average-- for discovering communities of standards by up to 90\%. 
Although these results required the validation of experts in the domain, an initial evaluation suggest that the results are accurate. 
%For the I40 domain, these results represent the possibility of verifying the experts knowledge. In addition, the possibility of unveiling new communities of standards based on their relations and not only on the classification of the standardization frameworks. 
%SemEP = 90.61
%METIS = 39.34
%KMeans= 85.66

\section{Related Work}
\label{sec:relatedwork}
In the literature, different approaches are proposed for discovering communities of standards as well as to corroborate and extend the knowledge of the standardization frameworks.
Zeid \emph{et al.}~\cite{zeid2019interoperability} study different approach to achieve interoperability of different standardization frameworks. 
In this work, the current landscape for smart manufacturing is described by highlighting the existing standardization frameworks in different regions of the globe. 
Lin \emph{et al.}~\cite{ramiirareport2017} present similarities and differences between the RAMI4.0 model and the IIRA architecture.
Based on the study of these similarities and differences authors proposed a functional alignment among layers in RAMI4.0 with the functional domains and crosscutting functions in IIRA. 
Monteiro \emph{et al.}~\cite{monteiro2018adoption} further report on the comparison of the RAMI4.0 and IIRA frameworks. 
In this work, a cooperation model is presented to align both standardization frameworks. 
Furthermore, mappings between RAMI4.0 IT Layers and the IIRA functional domain are established.
Another related approach is that outlined in~\cite{velasquez2018cloud}.
Moreover, the IIRA and RAMI4.0 frameworks are compared based on different features, e.g., country of origin, source organization, basic characteristics, application scope, and structure. 
It further details where correspondences exist between the IIRA viewpoints and RAMI4.0 layers.
Garofalo \emph{et al.}~\cite{abs-1808-00434} outline KGEs for I4.0 use cases. 
Existing techniques for generating embeddings on top of knowledge graphs are examined. 
Further, the analysis of how these techniques can be applied to the I4.0 domain is described; specifically, it identifies the predictive maintenance, quality control, and context-aware robots as the most promising areas to apply the combination of KGs with embeddings. 
All the approaches mentioned above are limited to describe and characterize existing knowledge in the domain. 
However, in our view, two directions need to be consider to enhance the knowledge in the domain; 1) the use of a KG based approach to encode the semantics; and 2) the use of machine learning techniques to discover and predict new communities of standards based on their relations. 

\section{Conclusion}
\label{sec:conclusion}
In this paper, we presented the  $\textit{I4.0}\cal{RD}$ approach that combines knowledge graphs and embeddings to discover associations between I4.0 standards. 
Our approach resorts to I4.0KG to discover relations between standards; I4.0KG represents relations between standards extracted from the literature or defined according to the classifications stated by the standardization frameworks. Since the relation between standards is symmetric and transitive, the transitive closure of the relations is materialized in I4.0KG. 
Different algorithms for generating embeddings are applied on the standards according to the relations represented in I4.0KG.
We employed three community detection algorithms, i.e., SemEP, METIS, and KMeans to identify similar standards, i.e., communities of standards, as well as to analyze their properties. Additionally, by applying the homophily prediction principle, novel relations between standards are discovered. We empirically evaluated the quality of the proposed techniques over 249 standards, initially related through 736 instances of the property \texttt{relatedTo}; as this relation is symmetric and transitive, its transitive closure is also represented in I4.0KG with 22,969 instances of \texttt{relatedTo}.
The Trans$^*$ family of embedding models were used to identify a low-dimensional representation of the standards according to the materialized instances of \texttt{relatedTo}. Results of a 5-fold cross validation process suggest that our approach is able to effectively identify novel relations between standards. Thus, our work broadens the repertoire of knowledge-driven frameworks for understanding I4.0 standards, and we hope that our outcomes facilitate the resolution of the existing interoperability issues in the I4.0 landscape.
As for the future work, we envision to have a more fine-grained description of the I4.0 standards, and evaluate hybrid-embeddings and other type of community detection methods.  

\bibliographystyle{splncs_srt}
\bibliography{BIBReferences}

\end{document}